\documentclass{article}
\usepackage{float}
\usepackage{amssymb}
\usepackage{booktabs}
\usepackage{graphicx}
\usepackage{subcaption}
\usepackage{enumitem}
\usepackage{bold-extra}
\usepackage{amsmath}
\usepackage{float}

\graphicspath{{figures/}}
\usepackage[preprint]{corl_2026} 

\title{Learning Motion Feasibility from Point Clouds in Cluttered Environments}

\author{
  Sajid Ansari, Arthi, Girish Varma, Antony Thomas\\
  International Institute of Information Technology Hyderabad\\
  Hyderabad 500032, India.\\
}

\begin{document}
\maketitle


\begin{abstract}
   Motion feasibility prediction plays a central role in robotics, particularly in task and motion planning and manipulation. A major bottleneck for this problem in cluttered environments is that infeasible planning attempts by Sampling-based motion planners (SBMPs) can incur substantial computational cost. Also existing approaches for infeasibility certification are limited to low-dimensional configuration spaces and often assume simplified geometric environments represented by primitive objects with known parameters. We study the complementary problem of learning motion feasibility prediction directly from raw RGB-D observations for a 7-DOF manipulator operating in realistic cluttered scenes. We introduce the first large-scale benchmark for this setting, comprising ~2.7M grasp feasibility labels over 88 scanned objects and 190 cluttered tabletop scenes. We benchmark three representative classifier families spanning MLP-based, volumetric-CNN, and point-cloud-based Transformer architectures under matched training conditions. Our best model, \textsc{GraspFC-PTX} (a point-cloud transformer), achieves an AUROC of 0.996 on Novel objects while providing predictions significantly faster than SBMPs.
\end{abstract}


\section{Introduction}
\label{sec:introduction}
Motion planning has been an active area of research for nearly five decades and remains a fundamental problem in robotics~\citep{orthey2023ARCRAS}. Despite substantial progress, complete motion planning, i.e, finding a collision-free path when one exists and certifying infeasibility otherwise, continues to be challenging~\citep{li2023IJRR}. Sampling-based motion planners (SBMPs)~\citep{karaman2011IJRR} have demonstrated remarkable scalability and can efficiently find feasible paths in high-dimensional spaces. However, in infeasible planning problems, i.e., when no collision-free path exists, SBMPs cannot provide certificates of infeasibility and typically continue searching until termination or timeout. Existing complete planning approaches are generally restricted to low-dimensional configuration spaces~\citep{zhang2008IJRR}, while recent advances in infeasibility certification remain limited to 5-DOF spaces~\citep{li2023RAL,thomas2025arxiv}. In this work, rather than explicitly certifying infeasibility, we address the complementary problem of learning a feasibility predictor that classifies planning instances as feasible or infeasible.

Determining whether a feasible motion exists is critical for many robotic planning frameworks. In task and motion planning (TAMP)~\cite{kaelbling2013IJRR,dantam2018IJRR,garrett2018IJRR,thomas2021RAS}, the feasibility of high-level task plans ultimately depends on the feasibility of the underlying motion planning problems. When a motion plan is infeasible, alternative task plans must be explored. \textit{However, in practice, infeasibility is often determined only after a planner reaches a predefined timeout, which can take several minutes and substantially increase overall planning time.} Feasibility evaluation is similarly central to manipulation in cluttered environments and rearrangement planning~\cite{stilman2007ICRA,dogar2011RSS,krontiris2015RSS,karami2021AIIA}. These problems frequently require identifying and relocating obstacles that obstruct task execution, where the need for obstacle displacement often arises only after a motion planning attempt is found to be infeasible. Likewise, in Navigation Among Movable Obstacles (NAMO)~\cite{stilman2005IJHR,muguiraIturralde2023ICRA}, the environment is actively modified by rearranging obstacles to create feasible paths. In all the above settings, the ability to rapidly predict infeasibility can substantially reduce planning overhead by avoiding expensive planning attempts that are unlikely to succeed, thereby enabling more efficient planning.

Recent infeasibility certification methods~\citep{li2023RAL,thomas2025arxiv} scale only to up to 5-DOF configuration spaces, whereas practical industrial manipulators typically operate in higher-dimensional spaces (e.g., 7-DOF). Moreover, these approaches assume environments composed of geometric primitives with known shape parameters. In contrast, real-world environments are commonly perceived as point clouds from RGB-D sensors. In this work, we directly operate on raw point cloud observations to predict motion feasibility. Our contributions are as follows:
\begin{itemize}[nosep]
    \item \textbf{Large-scale benchmark for motion feasibility prediction from real scenes.} We introduce a large-scale benchmark, built on top of GraspNet-1Billion~\citep{fang2020graspnet}, with 2.71M per-grasp RRT-Connect~\citep{kuffner2000ICRA} labels across 88 real scanned objects and 190 cluttered RGB-D tabletop scenes, including Seen/Similar/Novel splits and out-of-distribution settings (see Sections \ref{sec:dataset}, \ref{sec:ood}). The benchmark also evaluates motion feasibility of a 7-DOF manipulator directly from point clouds, eliminating the simplified geometric abstractions commonly used in prior infeasibility-prediction methods (Section \ref{sec:dataset}).
    \item \textbf{Systematic comparison across architecture families.} We benchmark three representative classifiers under matched data and compute budgets: (i) \textsc{GraspFC-NNet} (a MLP), (ii) \textsc{GraspFC-Conv3D} (a voxel-volumetric 3D-CNN), and (iii) \textsc{GraspFC-PTX} (a point-cloud transformer built on PTv3), all trained end-to-end from RRT-Connect supervision (Section \ref{sec:models}).
    \item \textbf{High accuracy and latency.} \textsc{GraspFC-PTX} achieves AUROC $0.996$ with TPR $98.5\,\%$ and TNR $97.1\,\%$ on the Novel objects, while all learned models provide predictions which is significantly faster then SBMPs (see Section \ref{sec:pred_results}). 
\end{itemize}

%
\section{Related Work}
\label{sec:related_work}
A growing body of work has explored learning-based techniques to accelerate or bypass the expensive motion planning stage~\citep{wells2019RAL, kim2019IJRR, silver2021AAAI, mcdonald2022CORL, bouhsain2023ICRA, bouhsain2024IROS, yang2023RSS}. Wells \textit{et al.}\citep{wells2019RAL} predicts geometric feasibility of symbolic actions using support vector machine trained on tabletop scenes involving two movable objects. However, feasibility is estimated through pairwise object evaluations which scales poorly with object count. Further, all objects are approximated as rectangular prisms. Neural network predicts entire symbolic action sequences from an RGB image and goal in~\citep{driess2021IJRR}. Training is performed on two-object scenes, with evaluation on up to six objects, considering only box-like and cylindrical object geometries. A graph neural network is employed to predict task-relevant object subsets (e.g., dishes, bowls, and cups in a dishwasher domain) in~\citep{silver2021AAAI}. However, the approach does not directly predict geometric feasibility and still requires motion planning for final feasibility verification. Kim \textit{et al.}~\citep{kim2019IJRR} predicts feasibility scores (e.g., low score for top grasp blocked by shelf) which helps prune infeasible or suboptimal plans early in the search process.

In~\citep{mcdonald2022CORL}, heuristic learning is coupled with learning policies that imitate the output of full TAMP planners, enabling the system to generate efficient action sequences by mimicking high-quality plans without repeatedly invoking expensive planning routines. However, this approach is limited to four-object pick and place tasks. Bouhsain \textit{et al.}~\citep{bouhsain2023ICRA} propose a neural network-based approach to predict the feasibility of discrete actions. Unlike earlier methods that primarily focus on single-object manipulation, their approach is capable of handling instances where multiple objects must be moved. This is enabled by a multi-channel image representation that captures different views of the scene, allowing the model to reason about object interactions more effectively. However, the network is trained only on scenes containing two objects, and its ability to generalize to more complex scenes with larger numbers of objects remains unvalidated. An extension of this approach to multi-robot scenarios is presented in~\cite{bouhsain2024IROS}. Yang \textit{et al.}~\citep{yang2023RSS} propose a Transformer-based model for task-plan feasibility prediction using object-level visual features extracted from segmented RGB images. The approach generates and ranks 100 candidate task plans based on predicted feasibility. However, generating a large number of candidate plans is computationally expensive.

Most existing approaches are trained on simplified environments with only a \textit{few} (typically two to four) movable objects and represent objects using simplified geometries such as \textit{cylinders} or \textit{box-shaped primitives}. Such assumptions can limit generalization and reduce prediction reliability in cluttered real-world environments with complex object geometries. In contrast, we use RGB-D point clouds of real cluttered tabletops with everyday objects of complex geometries to predict motion feasibility for a 7-DOF Franka Panda arm. 


We build on GraspNet-1Billion~\citep{fang2020graspnet} to annotate 2.71 million grasp candidates as feasible or infeasible. GraspNet-1Billion is a large-scale benchmark containing 97,280 RGB-D images with over one billion annotated grasp poses, along with an end-to-end grasp prediction network that takes point cloud inputs and predicts feasible grasp poses. However, the original benchmark does not account for whether a predicted grasp pose is kinematically reachable for a specific robot and its environment. To address this limitation, we augment the dataset by performing motion planning for every grasp candidate using a 7-DOF manipulator and assign binary reachability labels, categorizing grasps as either reachable or unreachable. This extension enables the study of motion-feasibility prediction in realistic manipulation scenarios where grasp quality alone is insufficient for successful execution. To the best of our knowledge, this is the first large-scale benchmark that provides per-grasp motion-feasibility annotations on real RGB-D scenes. An overview of the dataset construction pipeline is illustrated in Fig.~\ref{fig:method_overview}.

\section{Method}
\label{sec:approach}
\begin{figure*}[!t]
\centering
\includegraphics[trim=45 2 30 20,clip,scale=0.3]{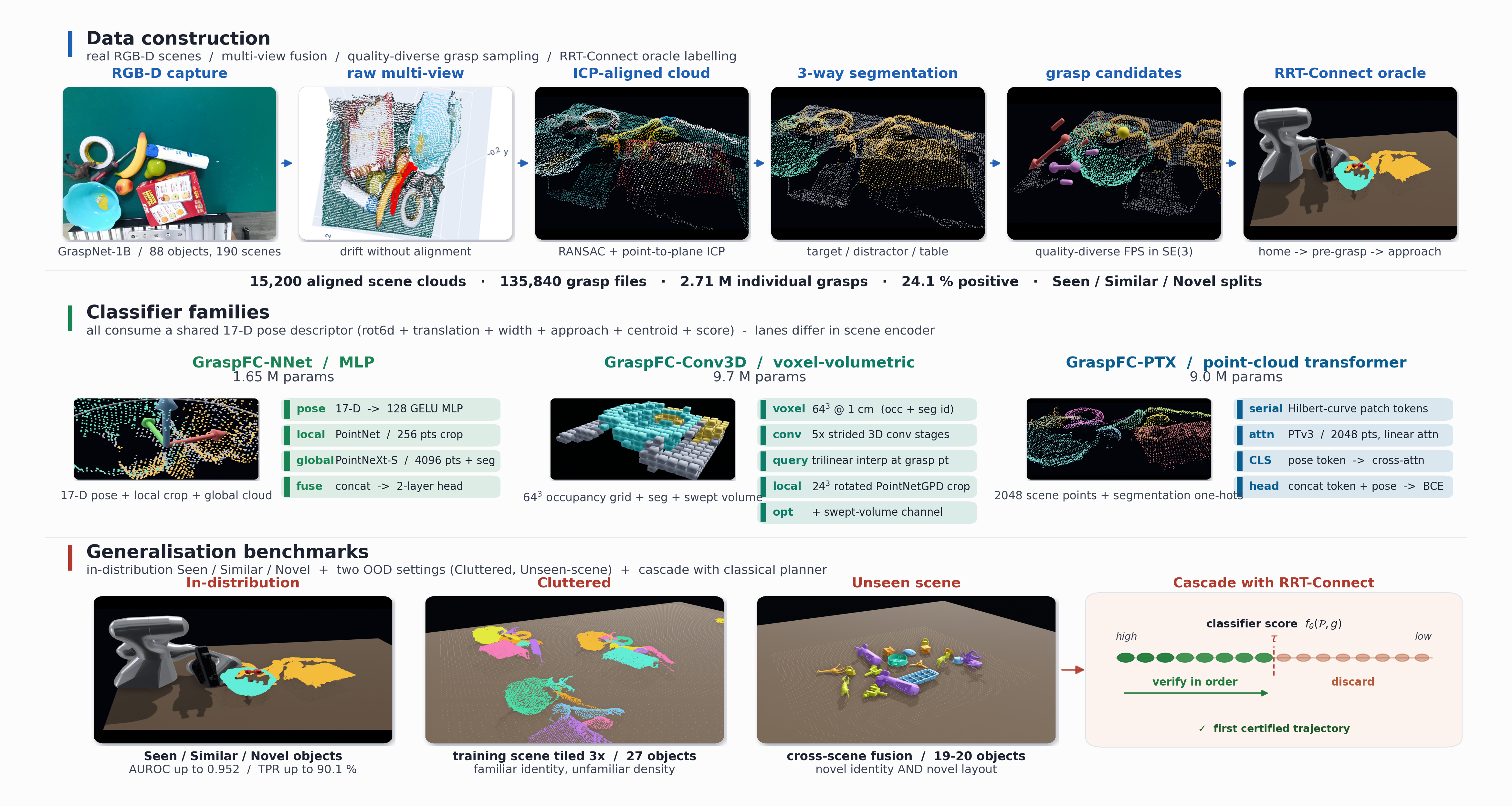}
\caption{
  Methodology overview.
  \textbf{Top}: data construction (Section~\ref{sec:dataset}) segmentation, quality-diverse grasp extraction, RRT-Connect labelling.
  \textbf{Middle}: the three classifier families compared (Section~\ref{sec:models}) GraspFC-NNet, GraspFC-Conv3D, GraspFC-PTX each sharing a 17-D pose descriptor but differing in scene representation.
  \textbf{Bottom}: evaluation on the in-distribution Seen/Similar/Novel splits and two out-of-distribution settings (Section~\ref{sec:ood_protocol}).}

\label{fig:method_overview}
\end{figure*}
We formulate motion feasibility prediction for grasping as a binary classification problem: \textit{given a partial scene observation and a candidate end-effector pose, predict whether a SBMP can find a collision-free trajectory from a given configuration to that pose}.
This section describes the data construction pipeline, three classifier families we benchmark, and their with a classical motion planner.
Fig.~\ref{fig:method_overview} summarizes the complete end-to-end pipeline.
\subsection{Problem Formulation}
\label{sec:problem}
Let $\mathcal{P} \in \mathbb{R}^{N \times 3}$ denote a scene point cloud expressed in the base frame of a manipulator $\mathcal{R}$ and $g = (R, t, w) \in SE(3) \times \mathbb{R}$ a parallel-jaw grasp with rotation $R$, translation $t$, and opening width $w$.
A grasp $g$ is \emph{feasible} in $\mathcal{P}$ if the end-effector of $\mathcal{R}$ can (i)~reach a pre-grasp standoff pose $p$ from the current configuration, and (ii)~execute a linear collision-free approach from $p$ to $g$. We use RRT-Connect in PyBullet~\citep{coumans2021} as the ground-truth oracle and learn a classifier $f_\theta(\mathcal{P}, g) \to [0,1]$ that predicts a binary verdict (feasible/infeasible) at a fraction of the computational cost.
Unlike prior feasibility predictors that operate on a small number of primitive-shaped  objects~\citep{wells2019RAL,driess2021IJRR,bouhsain2023ICRA}, our setting uses real RGB-D scenes containing scanned everyday objects of arbitrary geometry, requiring robust generalization across both grasp poses and scene compositions.

\subsection{Dataset Construction}
\label{sec:dataset}
\begin{figure}[t]
\centering
\includegraphics[width=\linewidth]{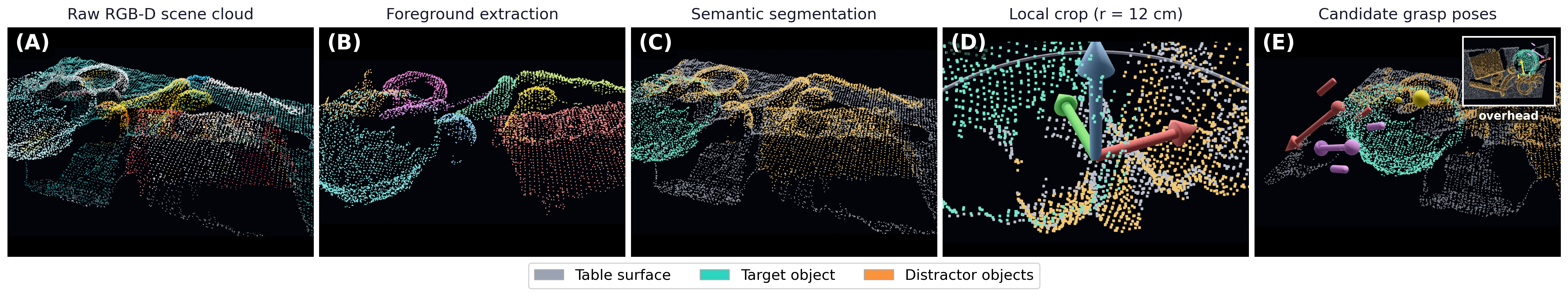}
\caption{%
  Data construction pipeline (left to right):
  (A) raw RGB-D scene cloud in the table frame;
  (B) foreground extraction with the background removed;
  (C) semantic segmentation into target, distractors, and table surface;
  (D) target-centric local crop in the grasp frame ($r = 12$\,cm);
  (E) candidate grasp poses with approach vectors and finger stubs (overhead view).}

\label{fig:pipeline}
\end{figure}

\paragraph{Aligned scene clouds.}
We build on GraspNet-1Billion~\citep{fang2020graspnet}, which contains 190 cluttered tabletop scenes captured using two depth cameras and 256 annotated frames per scene, spanning 88 objects with Seen, Similar, and Novel object splits.
For each scene, depth images are back-projected and transformed using the provided calibration. We align 40 sampled frames per (scene, camera) pair using RANSAC registration followed by point-to-plane ICP~\citep{besl1992pami}, producing 15{,}200 voxel-downsampled clouds with target, distractor, and table segmentation labels.
\paragraph{Valid grasp candidates.}
For each (scene, camera, frame, object) tuple we extract up to 20 grasp poses from the GraspNet 6-DoF annotations after applying force-closure and friction-based validity checks ($\mu \leq 0.4$).
To avoid spatial redundancy we perform a joint voxel--FPS downsampling that mixes grasp-quality ranking with farthest-point selection in $SE(3)$, retaining grasps that span the object's graspable surface rather than clustering on a single high-score region. The full set comprises 135{,}840 grasp files ($\approx$2.71M individual grasps).
\paragraph{Per-grasp labelling.}
Instead, \emph{Each individual grasp} is labelled with RRT-Connect as oracle, that includes (i)~Inverse Kinematics (IK) at the grasp pose, (ii)~IK at a 10\,cm pre-grasp standoff, (iii)~RRT-Connect from the home configuration to the pre-grasp pose with 10000 iterations, and (iv)~a 21-step linear approach check. The target object is retained in the obstacle cloud except for a small $\epsilon$-ball around the grasp point, giving realistic clutter without spuriously blocking the grasp itself.
The resulting per-grasp dataset is 24.1\% positive, a balance steep enough to be informative but not degenerate, and breaks down into four physically meaningful failure modes (IK, pre-grasp IK, RRT-connect path, approach collision) that we use for diagnostic analyses in Section~\ref{sec:result}. Fig.~\ref{fig:pipeline} illustrates the complete data construction pipeline.

\subsection{Classifier Families}
\label{sec:models}
We evaluate three representative architectures for 3D motion-feasibility prediction. The first, \textsc{GraspFC-NNet} (Motion Feasibility MLP), is a multilayer perceptron augmented with optional PointNeXt-S~\citep{qian2022pointnext} branches over the scene point cloud. The second, \textsc{GraspFC-Conv3D} (Motion Feasibility CNN), is a voxel-volumetric 3D-CNN that combines a GIGA-style trilinear feature query at the grasp location~\citep{jiang2021giga} with a PointNetGPD-style rotated local crop~\citep{liang2019pointnetgpd}. The third, \textsc{GraspFC-PTX} (Motion Feasibility Point Cloud Transformer), is built on a Point Transformer V3 (PTv3) backbone~\citep{wu2024ptv3}, into which the grasp pose is injected through a single classification token, complemented by register tokens~\citep{darcet2024iclr} and a target-aware attention bias. To our knowledge, \textsc{GraspFC-PTX} is the first application of PTv3 to per-grasp motion feasibility prediction, and the first head-to-head comparison of MLP-based, voxel-volumetric, and transformer-based feasibility predictors on real RGB-D point clouds.

All three models take the same 17-D pose descriptor, comprising 6D continuous rotation~\citep{zhou2019rot6d}, translation, gripper width, unit approach vector, target-object centroid, and GraspNet quality score, and differ mainly in how scene geometry is represented. To our knowledge, this is the first application of PTv3 to per-grasp motion-feasibility prediction, and the first systematic comparison of MLP-based, voxel-volumetric, and transformer-based feasibility predictors on real RGB-D point clouds.

\noindent\textbf{\textsc{GraspFC-NNet} (1.65M params).}
The 17-D grasp descriptor is processed by a 128-unit MLP head with LayerNorm and GELU activations and augmented with two scene-geometry branches: (i) a PointNet~\citep{qi2017pointnet} over a 256-point local crop in the grasp frame ($r=12$\,cm), and (ii) a global PointNeXt-S~\citep{qian2022pointnext} over 4{,}096 FPS-sampled scene points with three-way segmentation one-hots.
The three feature streams are concatenated and fused with a two-layer MLP, with branch-design.

\noindent \textbf{\textsc{GraspFC-Conv3D} (9.7M params).} The model follows the volumetric backbone of VGN~\citep{breyer2021vgn} and GIGA~\citep{jiang2021giga}, processing a $64^3$ occupancy grid at 1 cm resolution (see Fig.~\ref{fig:app_inputs_1}, Appendix~\ref{sec:appendix}) with five strided 3D convolution layers. Scene features are queried at the grasp location using trilinear interpolation in the feature volume, following GIGA. A PointNetGPD-style~\citep{liang2019pointnetgpd} rotated local crop of size $24^3$ (8 cm cube in the grasp frame) is processed by a lightweight auxiliary 3D CNN. The queried grid feature, local feature, and pose embedding are fused at the classification head. We optionally add a binary voxel channel encoding the gripper swept volume along the approach direction to evaluate whether explicit path geometry improves feasibility prediction.


\noindent \textbf{\textsc{GraspFC-PTX} (9.0M params).}
Point Transformer V3~\citep{wu2024ptv3} encodes 2{,}048 scene points (segmentation one-hots as input features) via Hilbert-curve serialised patch attention with linear-time complexity. 
We inject the grasp pose through a learned classification (CLS) token (see Fig.~\ref{fig:app_inputs_2}, Appendix~\ref{sec:appendix}) whose initial embedding is the output of a 128-unit pose MLP added to a learnable token initialisation; the CLS token cross-attends to PTv3's encoder tokens, and the cross-attention output is concatenated with the pose embedding before the BCE head.
When the swept-volume channel is enabled, it is concatenated as a fifth per-point feature.
PTv3 lets us test whether a state-of-the-art point-cloud backbone narrows the gap to the much smaller MLP when scene geometry actually carries signal.

\subsection{Out-of-Distribution Evaluation}
\label{sec:ood_protocol}
The in-distribution benchmark draws train and test scenes from the same single-tabletop GraspNet distribution.
To evaluate whether the cascade transfers to scenes that no model in the literature has been trained on, we construct two complementary out-of-distribution (OOD) settings.

The first, \emph{Cluttered}, tiles a single held-out training scene three times along the workspace to yield 27 objects of familiar identity but unfamiliar density.
The second, \emph{Unseen scene}, fuses objects sampled from three different held-out Novel scenes into a single tabletop 19--20 objects whose identity \emph{and} layout are unseen during training.

For both the OOD settings we run the full data pipeline (alignment, grasp extraction, RRT-Connect labelling) from scratch with the same parameters as the in-distribution benchmark and report mean$\pm$std over three random seeds per scene to capture the stochasticity of point-cloud sub-sampling at inference time.
\section{Experimental Results}
\label{sec:result}

\begin{figure}[t]
\centering
\includegraphics[width=\linewidth]{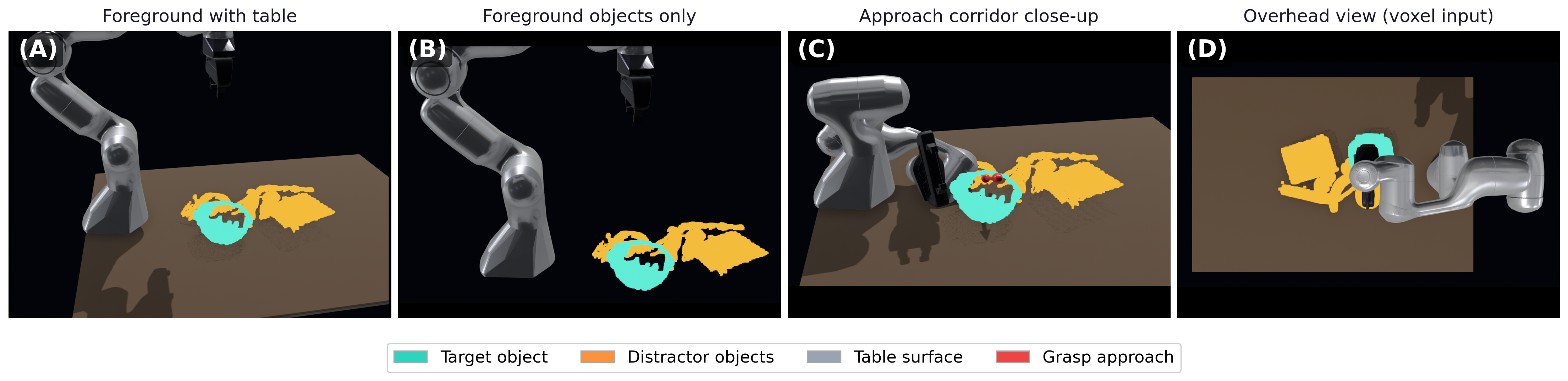}
\caption{%
  Scene representations used as classifier inputs.
  (A) point cloud with table mesh;
  (B) foreground objects with the table removed;
  (C) grasp-aligned close-up showing the gripper's approach corridor;
  (D) top-view of the scene.
  Colours: target object (teal), distractor objects (orange), table surface (grey), grasp approach arrow (red).
}
\label{fig:scene_variants}
\end{figure}

The dataset constructed in Section~\ref{sec:dataset} is evaluated on the three standard GraspNet-1Billion splits: \textbf{Seen}, \textbf{Similar}, and \textbf{Novel} (30 scenes each), with 50{,}000 samples per split at the natural class distribution. Fig.\ref{fig:scene_variants} shows the four classifier inputs. We consider two training subsets: \textsc{MoFeas-5K-Bal} (5{,}000 quality-filtered, class-balanced grasps) and \textsc{MoFeas-200K} (200{,}000 grasps at the natural 24.1\% positive rate). Further details regarding dataset statistics and parameter counts/input variants of models are reported in Appendix~\ref{app:dataset} and \ref{app:configs}.



The models \textsc{GraspFC-NNet}, \textsc{GraspFC-Conv3D}, and \textsc{GraspFC-PTX} are trained \emph{end-to-end from random initialisation} on our feasibility data without any pretrained weights.

%
\subsection{Feasibility Prediction Results}
\label{sec:pred_results}

Table~\ref{tab:main} reports the in-distribution results on Seen, Similar, and Novel.


\noindent\textbf{Architecture convergence on \textsc{MoFeas-5K-Bal}.}
At this data size, all three models land within 0.015 AUROC of one another (\textsc{GraspFC-NNet}: 0.981, \textsc{GraspFC-Conv3D}: 0.967, \textsc{GraspFC-PTX}: 0.982).
The lightest model matches the 9M-parameter transformer: stronger structural assumptions built into larger architectures provide no systematic benefit when data is the bottleneck. Ablations on pose injection and attention mechanisms are provided in Appendix~\ref{app:ptv3}.

\noindent\textbf{Data scale is the dominant lever.}
Scaling from \textsc{MoFeas-5K-Bal} to \textsc{MoFeas-200K} yields $+0.014$--$0.015$ AUROC across all models  matching or exceeding the gain from any architectural change.
The best model, \textsc{GraspFC-PTX} trained on \textsc{MoFeas-200K}, achieves \textsc{AUROC 0.9956} on Novel with TPR 98.5\,\% and TNR 97.1\,\%. More details can be found in Appendix~\ref{app:scale}.

\noindent\textbf{Robust cross-category generalisation.}
For every model in Table~\ref{tab:main}, the AUROC gap between Seen and Novel is $<0.002$, indicating reliable transfer to the 30 unseen object geometries.
This near-zero gap across architectures and scales suggests that models learn a generalised kinematic rule rather than object-specific appearance shortcuts.

\begin{table*}[!t]
\centering
\caption{%
  Feasibility classification on GraspNet-1Billion.
  AUROC on the three splits; TPR and TNR on \textbf{Novel} split.
  Inputs (\emph{Pose}, \emph{Cloud}, \emph{Voxel}) are defined in Section~\ref{sec:models}; backbones cited: VGN~\citep{breyer2021vgn}, GIGA~\citep{jiang2021giga}, PointNeXt-S~\citep{qian2022pointnext}, PTv3~\citep{wu2024ptv3}.
}
\label{tab:main}
\footnotesize
\setlength{\tabcolsep}{4pt}
\resizebox{\textwidth}{!}{%
\begin{tabular}{lllc @{\hspace{4pt}} ccc @{\hspace{4pt}} cc}
\toprule
 & & & & \multicolumn{3}{c}{AUROC} & \multicolumn{2}{c}{Novel metrics} \\
\cmidrule(lr){5-7}\cmidrule(lr){8-9}
Model & Encoder (Architecture) & Input & Train. set & Seen & Similar & Novel & TPR\,(\%) & TNR\,(\%) \\
\midrule
\multicolumn{9}{l}{\textit{Scene-only control (no grasp pose given to the model)}} \\
\textsc{GraspFC-Conv3D} (no pose) & VGN          & Scene cloud  & \textsc{MoFeas}-5K-Bal & 0.552 & 0.565 & 0.521 & 42.3 & 61.3 \\
\midrule
\textsc{GraspFC-Conv3D}  & VGN + GIGA   & Pose + Voxel & \textsc{MoFeas}-5K-Bal & 0.964 & 0.964 & 0.967 & 91.1 & 92.3 \\
\textsc{GraspFC-PTX}     & PTv3         & Pose + Cloud & \textsc{MoFeas}-5K-Bal & 0.981 & 0.981 & 0.982 & 89.1 & 96.2 \\
\textsc{GraspFC-Conv3D}  & VGN + GIGA   & Pose + Voxel & \textsc{MoFeas}-200K   & 0.980 & 0.978 & 0.981 & 97.4 & 91.0 \\
\textsc{GraspFC-NNet}    & PointNeXt-S  & Pose + Cloud & \textsc{MoFeas}-200K   & 0.995 & 0.995 & 0.995 & 97.9 & 96.3 \\
\textsc{GraspFC-PTX}     & PTv3         & Pose + Cloud & \textsc{MoFeas}-200K   & \textbf{0.996} & \textbf{0.996} & \textbf{0.996} & \textbf{98.5} & \textbf{97.1} \\
\bottomrule
\end{tabular}%
}
\end{table*}

\subsection{Runtime Comparisons for Planning}
\label{sec:speedup}

We compare per-query decision time on $5{,}910$ Novel grasps for the RRT-Connect planner and the three classifiers, as summarized in Table~\ref{tab:latency}.
\emph{Per-query latency} is the wall-clock to a single feasibility verdict for one (scene, grasp) pair.
For the planner this is the time to either return a path or declare infeasibility at the search budget; we report the mean separately for feasible and infeasible queries because the two distributions differ by an order of magnitude.
For the classifier, latency is the forward pass on GPU at batch one and is constant in the true label. Since the infeasible case is the worst case that learned filtering is meant to amortise, the ratio of the mean infeasible-planner cost ($754\,\text{ms}$) to the classifier latency is reported using the \emph{Speedup} metric.

Given the tabletop setting and 7-DOF arm, we use 10000 iterations for the the planner.\footnote{We run RRT-Connect with a 10000-iteration limit and label grasps that fail to return a solution within this budget as infeasible.}
The planner takes $11\times$ longer on infeasible than feasible queries: a feasible grasp typically terminates the bidirectional search early ($12\,\text{ms}$ median), whereas an infeasible grasp drives the planner to exhaust its sampling budget ($p95\ 1.92\,\text{s}$).
Every learned classifier predicts in under $10\,\text{ms}$, giving $84\text{--}171\times$ speedup over the infeasible-planner mean.
\textsc{GraspFC-NNet} at $4.4\,\text{ms}$ is the cheapest predictor; \textsc{GraspFC-PTX} adds $3\,\text{ms}$ but is the most accurate under distribution shift (Section~\ref{sec:ood}).
The asymmetry has a practical consequence: pre-filtering with the classifier is most beneficial precisely on the candidates the planner is slowest to reject, which is the case the cascade in Appendix~\ref{sec:cascade} exploits.

\begin{table}[!htbp]
\centering
\caption{%
  Per-query decision time on Novel grasps with Speedup ratio.
}
\label{tab:latency}
\footnotesize
\setlength{\tabcolsep}{4pt}
\resizebox{\linewidth}{!}{%
\begin{tabular}{llcc}
\toprule
Method & Variant & Latency / query & Speedup vs.\ infeasible planner \\
\midrule
Motion planner & Feasible grasp   & 67\,ms (mean), 12\,ms (median)  & $11.3\times$ \\
Motion planner & Infeasible grasp & 754\,ms (mean), 1.92\,s (p95)   & $1\times$ (reference) \\
\midrule
\textsc{GraspFC-NNet}   & Pose + scene cloud (1.65M params)  & 4.4\,ms                           & $171\times$ \\
\textsc{GraspFC-Conv3D} & Pose + voxel (9.7M params)         & 9.0\,ms                           & $84\times$  \\
\textsc{GraspFC-PTX}    & Pose + scene cloud (9M params)     & 7.4\,ms                           & $102\times$ \\
\bottomrule
\end{tabular}%
}
\end{table}

\subsection{Out-of-Distribution Results}
\label{sec:ood}
\begin{figure}[!htbp]
\centering
\begin{subfigure}[t]{0.24\linewidth}
  \centering
  \includegraphics[width=\linewidth]{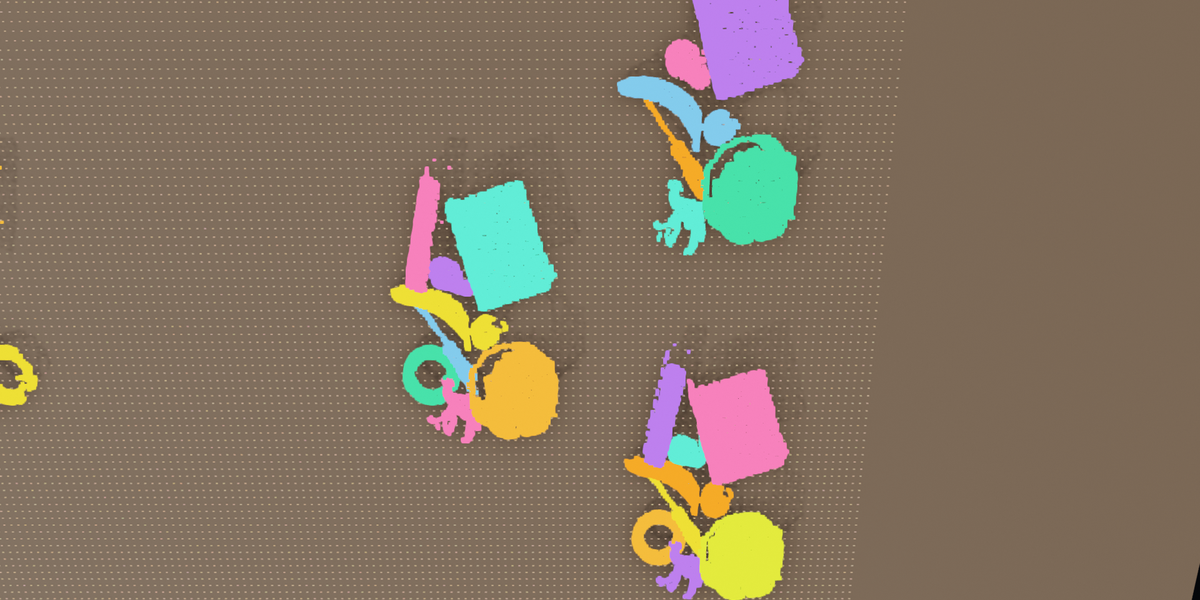}
  \caption{Cluttered (top-down).}
\end{subfigure}\hfill
\begin{subfigure}[t]{0.24\linewidth}
  \centering
  \includegraphics[width=\linewidth]{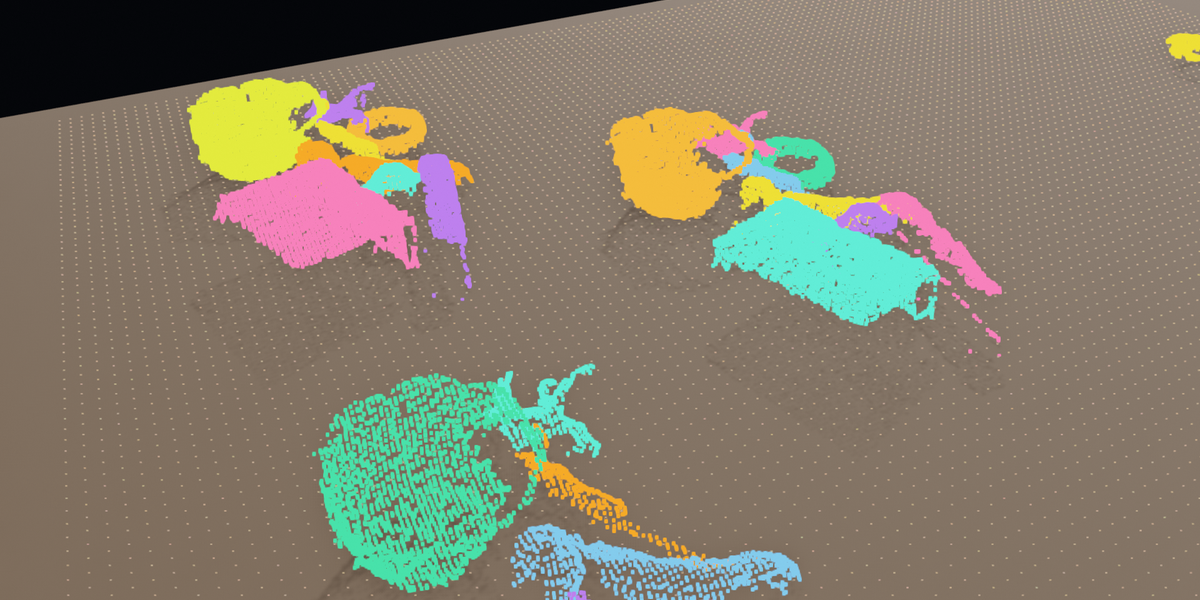}
  \caption{Cluttered (close-up).}
\end{subfigure}\hfill
\begin{subfigure}[t]{0.24\linewidth}
  \centering
  \includegraphics[width=\linewidth]{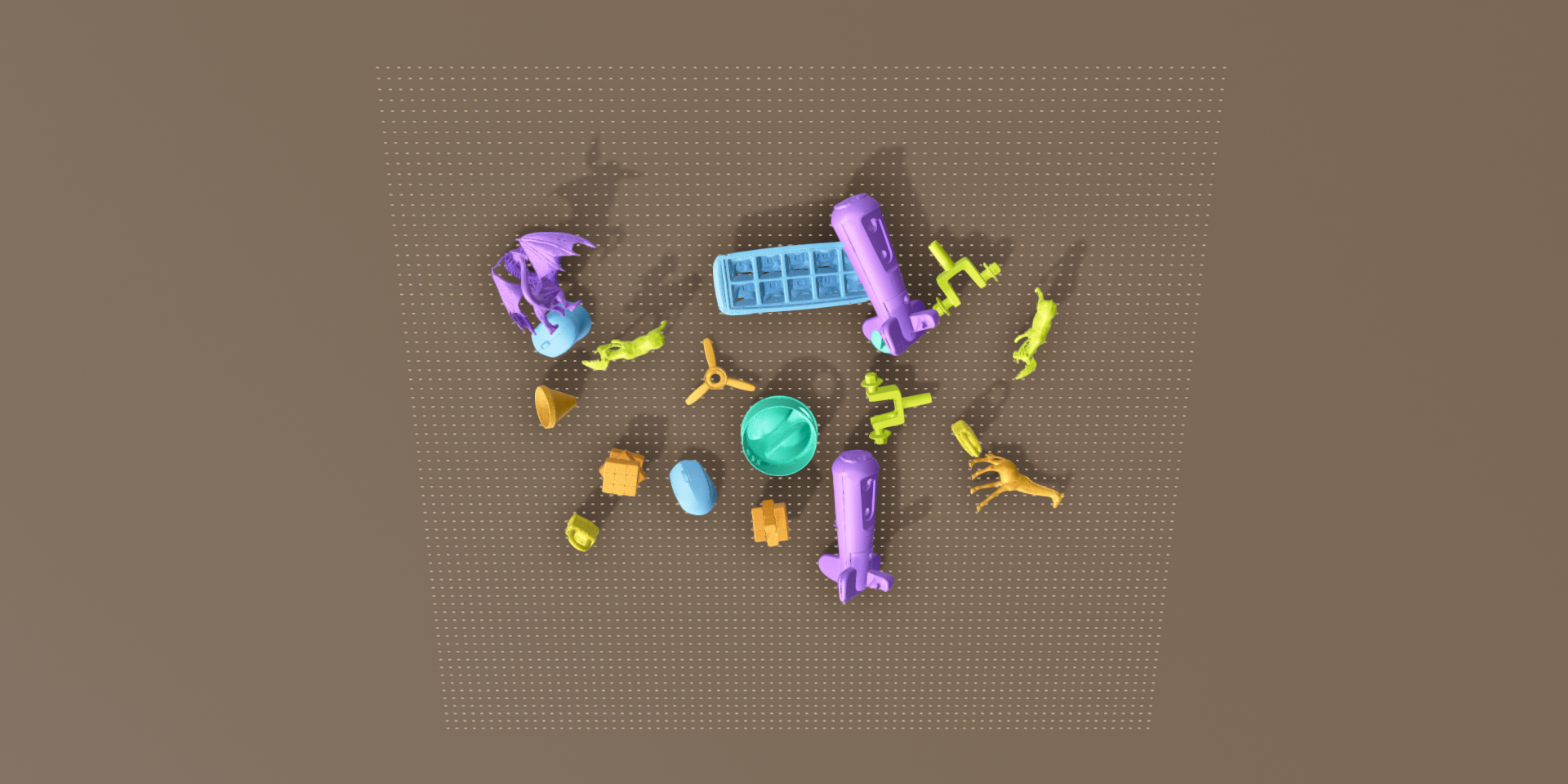}
  \caption{Unseen scene (top-down).}
\end{subfigure}\hfill
\begin{subfigure}[t]{0.24\linewidth}
  \centering
  \includegraphics[width=\linewidth]{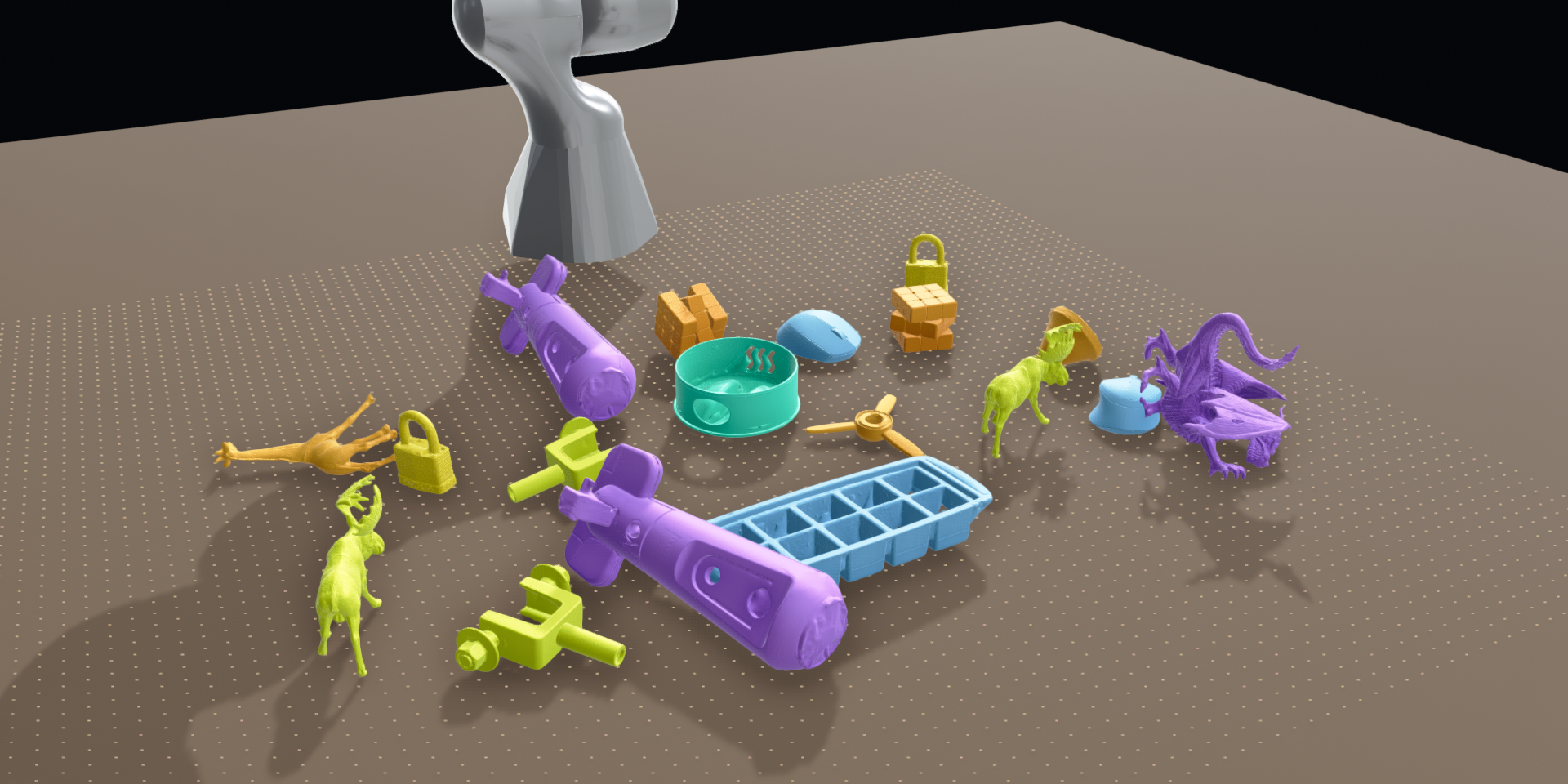}
  \caption{Unseen scene (close-up).}
\end{subfigure}
\caption{%
  Out-of-distribution settings, top-down view. (a,b) \emph{Cluttered} and (c,d) \emph{Unseen scene}. 
}
\label{fig:ood_settings}
\end{figure}

We follow the two OOD protocols defined in Section~\ref{sec:ood_protocol} and report results on three random seeds for the Unseen and Cluttered scenes, as shown in Fig.~\ref{fig:ood_settings}.

We report two complementary metrics in Table~\ref{tab:ood}: classifier AUROC (threshold-free rank quality) and \emph{Top-$k$ grasp success} which is the probability that at least one of the classifier's $k$ highest-scored candidates is verified feasible by the planner.
Top-$k$ success directly measures the downstream utility for a TAMP system that attempts grasps in score order and stops at the first executable trajectory.

\begin{table}[!htbp]
\centering
\caption{%
  Out-of-distribution evaluation. Trained on \textsc{MoFeas}-200K; mean $\pm$ std over 3 scenes $\times$ 3 seeds per setting.
}
\label{tab:ood}
\small
\begin{tabular}{llrrrr}
\toprule
Model & Setting & AUROC & TPR\,(\%) & TNR\,(\%) & Top-5\,(\%) \\
\midrule
\textsc{GraspFC-NNet}   & Cluttered    & $0.977 \pm 0.009$          & $87.7 \pm 4.4$          & $96.0 \pm 2.1$          & $100$ \\
\textsc{GraspFC-Conv3D} & Cluttered    & $0.903 \pm 0.022$          & $82.3 \pm 2.7$          & $89.0 \pm 2.7$          & $100$ \\
\textsc{GraspFC-PTX}    & Cluttered    & $\mathbf{0.993 \pm 0.003}$ & $\mathbf{96.7 \pm 3.1}$ & $\mathbf{97.2 \pm 1.3}$ & $100$ \\
\midrule
\textsc{GraspFC-NNet}   & Unseen scene & $0.846 \pm 0.014$          & $74.9 \pm 1.9$          & $85.5 \pm 3.4$          & $100$ \\
\textsc{GraspFC-Conv3D} & Unseen scene & $0.798 \pm 0.008$          & $69.6 \pm 2.2$          & $82.7 \pm 2.9$          & $100$ \\
\textsc{GraspFC-PTX}    & Unseen scene & $\mathbf{0.852 \pm 0.021}$ & $\mathbf{81.1 \pm 1.6}$ & $\mathbf{88.2 \pm 2.4}$ & $100$ \\
\bottomrule
\end{tabular}
\end{table}

\noindent\textbf{Graceful degradation where it matters.}
The cluttered setting incurs only a modest AUROC drop ($-0.003$ to $-0.10$ relative to in-distribution); the harder unseen-scene setting costs $0.14$--$0.20$ AUROC.
\textsc{GraspFC-PTX} degrades least in both regimes, consistent with its larger inductive capacity for scene geometry.
Crucially, \emph{Top-5 grasp success is $100\,\%$ across all $18$ (scene, seed) combinations}: although score \emph{calibration} suffers under distribution shift, the relative \emph{ranking} of candidates is preserved well enough that a feasible grasp is consistently among the top five.
For the downstream cascade  classifier ranking followed by planner verification  this is the relevant operating regime, and the learned predictors transfer to cluttered and unseen scenes without retraining.

\noindent\textbf{Planner cost amplifies under distribution shift.}
The planner's worst case  attempting an infeasible grasp until the search budget is exhausted  remains expensive in OOD scenes: mean infeasible-grasp planning time is $1.11\,$s in the cluttered setting and $1.09\,$s in the unseen-scene setting (Table~\ref{tab:ood_timing}).
Classifier latency is invariant to scene complexity, so the cost ratio between the planner and any classifier stays in the $100\times$--$1000\times$ range across all OOD scenes, with the \textsc{GraspFC-NNet} exceeding $1000\times$ on the cluttered setting.
The cascade architecture remains highly beneficial as scenes drift away from the training distribution.

\begin{table}[H]
\centering
\caption{%
  Per-query timing under distribution shift (mean $\pm$ std over 3 scenes $\times$ 3 seeds per setting).
  In-distribution numbers from Table~\ref{tab:latency}.
}
\label{tab:ood_timing}
\footnotesize
\setlength{\tabcolsep}{4pt}
\resizebox{\linewidth}{!}{%
\begin{tabular}{lccccc}
\toprule
              & Planner            & Planner               & \multicolumn{3}{c}{Cost ratio (planner-infeas / classifier)} \\
\cmidrule(lr){4-6}
Setting       & Feasible (mean)    & Infeasible (mean)     & \textsc{GraspFC-NNet} & \textsc{GraspFC-Conv3D} & \textsc{GraspFC-PTX} \\
\midrule
In-distribution (Novel)      & $67\,$ms          & $754\,$ms             & $171\times$            & $84\times$            & $102\times$ \\
Cluttered                    & $98\,$ms          & $1.11 \pm 0.03\,$s    & $\mathbf{761 \pm 289}\times$ & $112 \pm 10\times$ & $175 \pm 5\times$ \\
Unseen scene                 & $46\,$ms          & $1.09 \pm 0.41\,$s    & $339 \pm 53\times$     & $133 \pm 31\times$    & $\mathbf{194 \pm 28}\times$ \\
\bottomrule
\end{tabular}%
}
\end{table}

\section{Limitation and Conclusion}
\label{sec:limitation}
\textbf{Un-seen scene feasibility accuracy drops.} 
Though we achieve high AUROC of 99\% in in-distribution and Cluttered OOD settings, it drops to 85\% on the Unseen-scene OOD setting, where both object identity and layout differ from training (see Table \ref{tab:ood}).
Top-5 grasp success stays at 100\,\%, but closing this gap likely requires training on a cross-scene mixture rather than the single-scene-per-table regime of GraspNet-1Billion.

\textbf{Limited grasp poses per-target object and number of objects per scene.} Currently, we sample 20 grasp poses for each target object in the scene. Further, Graspnet-1Billion contains an average of only~9 objects per scene. While this is sufficient for moderately cluttered environments, heavily cluttered scenes with densely packed objects may require a substantially larger number of grasp samples to ensure adequate coverage and balanced labels for each object. In such scenarios, the probability of infeasible grasp poses increases significantly due to occlusions and kinematic constraints, making dense grasp sampling essential for reliable motion-feasibility learning. 

We introduced the first large-scale point-cloud motion-feasibility benchmark for a 7-DoF arm, 2.71M per-grasp RRT-Connect labels over 88 real scanned objects and used it to compare \textsc{GraspFC-NNet}, \textsc{GraspFC-Conv3D}, and \textsc{GraspFC-PTX} under matched data and compute.
\textsc{GraspFC-PTX} trained on \textsc{MoFeas-200K} reaches AUROC 0.996 with 98.5\,\% TPR and 97.1\,\% TNR on the Novel-object split and, when composed with RRT-Connect as a pre-filter and score-based ranker, delivers $80$--$200\times$ speedup over the planner's worst case while preserving its completeness guarantee.




\newpage
%


\bibliography{references}  

\clearpage
\appendix
\section{Extended Ablations and Design Studies}
\label{sec:appendix}

\begin{figure}[H]
\centering
\begin{subfigure}[t]{0.31\linewidth}
  \centering
  \includegraphics[width=\linewidth]{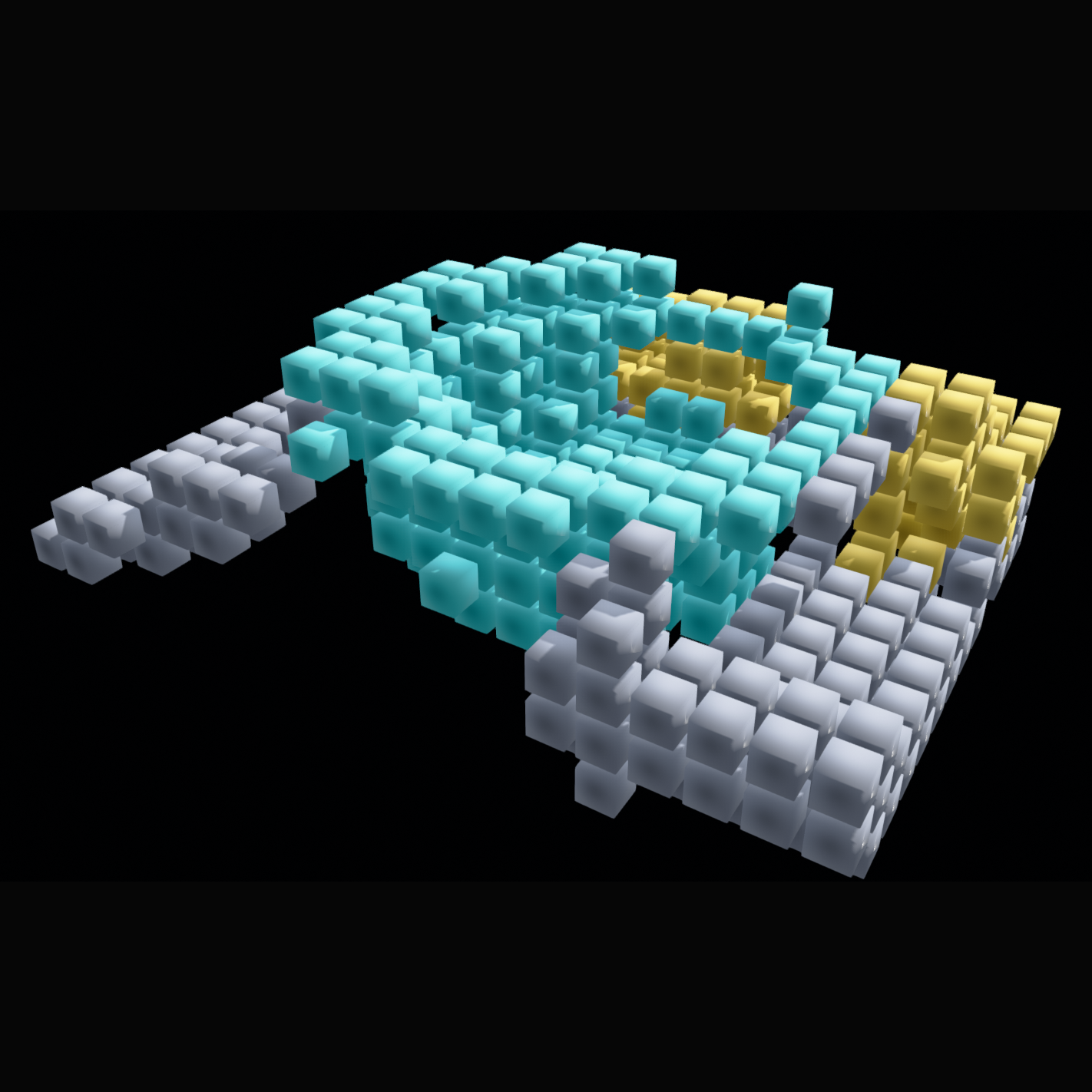}
  \caption{$64^3$ voxel grid consumed by GraspFC-Conv3D.}
  \label{fig:app_inputs_1}
\end{subfigure}\hfill
\begin{subfigure}[t]{0.31\linewidth}
  \centering
  \includegraphics[width=\linewidth]{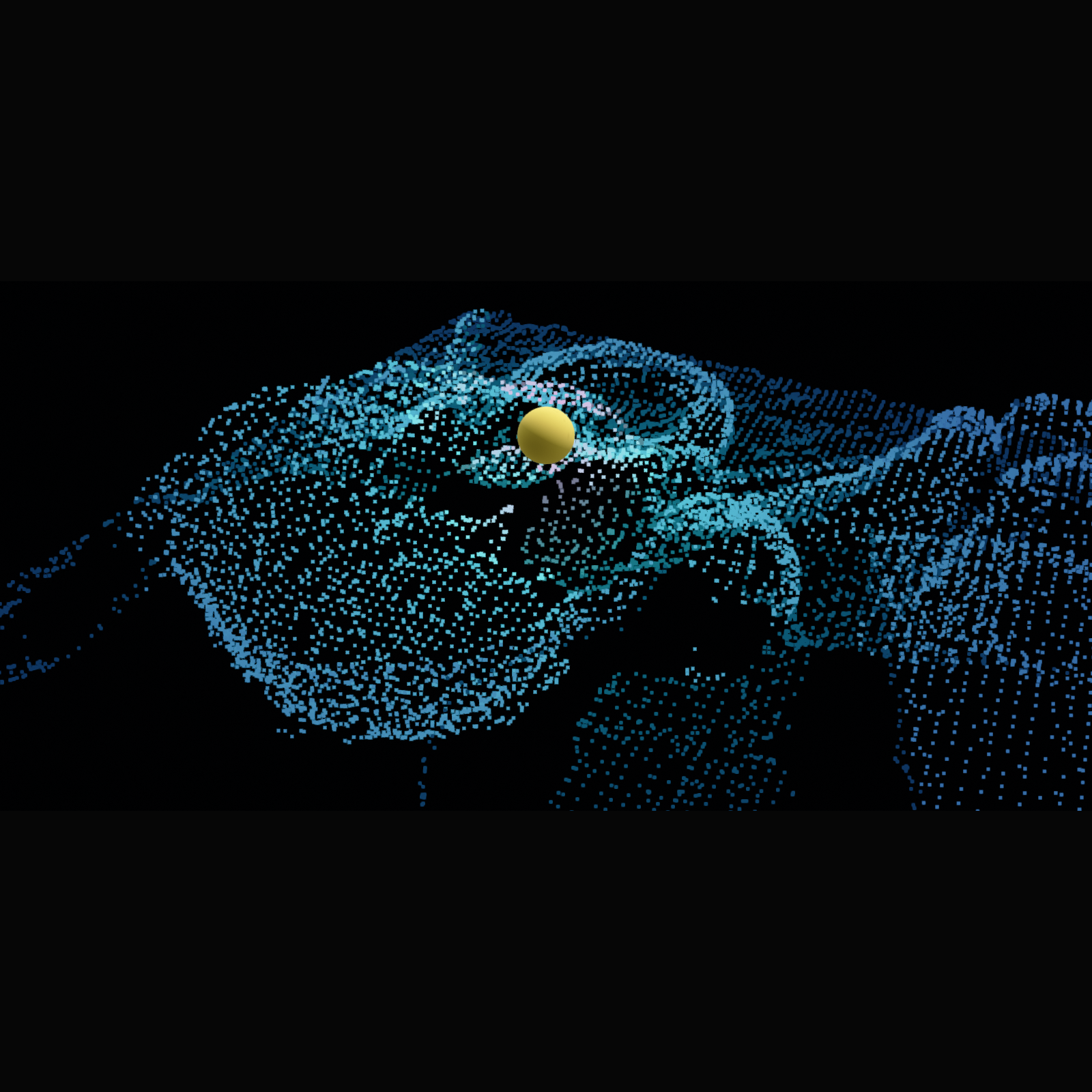}
  \caption{GraspFC-PTX CLS-token attention concentrating on the target object.}
  \label{fig:app_inputs_2}
\end{subfigure}\hfill
\begin{subfigure}[t]{0.31\linewidth}
  \centering
  \includegraphics[width=\linewidth]{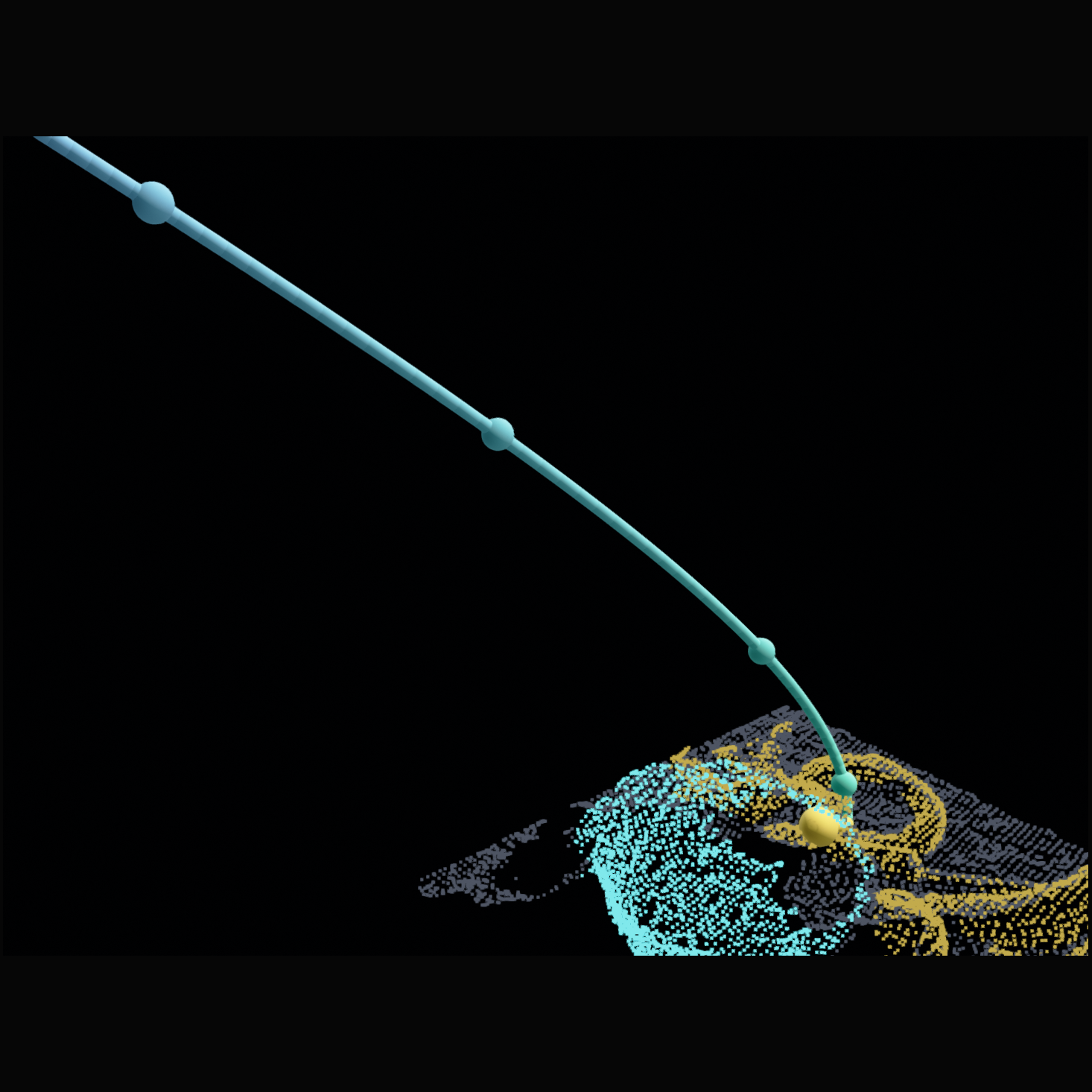}
  \caption{RRT-Connect trajectory used to label each grasp.}
\end{subfigure}
\caption{Model-side representations consumed by each architecture and the planner-verified trajectory used to label each grasp.}
\label{fig:app_inputs}
\end{figure}

This appendix details the design studies behind the main-paper results.
Each table reports AUROC, TPR and TNR on the Novel split with the natural $24\,\%$-positive class distribution.
Tables~\ref{tab:app_dataset}--\ref{tab:app_classbal} use the same labels and protocols as Section~\ref{sec:result}.
Fig.~\ref{fig:app_inputs} visualises the three model-side inputs that the families consume.

\subsection{Training Details}
All three families are trained end-to-end from random initialisation with BCE-with-logits loss, AdamW ($\mathrm{lr}{=}1.5{\times}10^{-3}$, cosine decay), bf16 mixed precision, and a positive-class weight $w_{+} \approx 3.14$ computed from the natural 24.1\% positive rate.
The decision threshold is tuned on a held-out validation split by maximising balanced accuracy and is frozen for all test evaluations.
On \textsc{MoFeas-200K}, the natural-distribution training subset used for our main results, each architecture converges within 30--50 epochs on a single RTX~4090/5090.

\subsection{Planner Integration: Cascade Evaluation}
\label{sec:cascade}

A learned classifier alone is not a planner; it cannot produce executable trajectories.
We therefore integrate $f_\theta$ with RRT-Connect as a \emph{cascade} that preserves the planner's correctness guarantees while exploiting the classifier's constant-time predictions to amortise the planner's long worst case on infeasible grasps (mean 754\,ms, p95 1.92\,s).
Two cascade modes serve different downstream needs:

\noindent\emph{Pre-filtering}~\citep{driess2022nfc}: candidates with $f_\theta(\mathcal{P}, g) < \tau$ are discarded; the planner is invoked only on those above the threshold $\tau$, eliminating the bulk of infeasible attempts where the planner is slowest.

\noindent\emph{Score-based ranking}~\citep{yang2023RSS}:
: the planner verifies candidates in decreasing classifier-score order and stops at the first feasible trajectory; no candidate is dropped, so feasible recall is unchanged.

The two modes compose: pre-filtering removes the infeasible mass, and ranking the survivors minimises time to the first certified solution.
Both rely on the classifier's score \emph{ordering} rather than its absolute calibration, which makes the cascade robust to the calibration drift we observe under distribution shift (Section~\ref{sec:ood}).

The advantage is \emph{asymmetric}: on easy feasible grasps the planner's early-stopping (median 12\,ms) is faster than the classifier's fixed overhead.
We therefore adopt a \emph{cascade evaluation}: the classifier scores all candidates in one batch, and the motion planner is invoked only on candidates retained by the cascade to obtain verified, executable trajectories.
At a batch of 64, the classifier screens 430 candidates per second versus the planner's effective ${\sim}1.3$ grasps/s.

Tables~\ref{tab:prefilter} and~\ref{tab:rank} quantify the two standard cascade modes evaluated on Novel.

\begin{table}[!htbp]
\centering
\caption{%
  Pre-filtering on Novel (5{,}568 candidates, 24.6\,\% feasible). Total time = aggregated planner latency on the queried subset.
}
\label{tab:prefilter}
\small
\begin{tabular}{lcccc}
\toprule
Configuration & Planner calls & Total planner time & Speedup & Feasible retained \\
\midrule
No pre-filter (baseline) & 5,568 & 5,470\,s          & $1\times$             & 100\,\%  \\
With pre-filter          & 1,743 & \phantom{0,}600\,s & $\mathbf{9.1\times}$ & 92.3\,\% \\
\bottomrule
\end{tabular}
\end{table}

\begin{table}[!htbp]
\centering
\caption{%
  Score-based ranking on Novel. Time = mean wall-clock to the first planner-verified feasible grasp.
}
\label{tab:rank}
\small
\begin{tabular}{lcccc}
\toprule
Configuration & Order & Time to first feasible & Speedup & Top-1 accuracy \\
\midrule
No ranking (baseline) & random scan & 820\,ms & $1\times$             & --- \\
With ranking          & by score    & \textbf{201\,ms} & $\mathbf{4.1\times}$ & 94.8\,\% \\
\bottomrule
\end{tabular}
\end{table}

Pre-filtering yields the largest speedup ($9.1\times$, 89\,\% planner time saved) at a cost of 7.7\,\% feasible-grasp recall, adjustable by the classifier threshold.
Score-based ranking retains every candidate and reduces time-to-first-solution by $4.1\times$ at zero recall cost.
The two modes are complementary: pre-filtering removes the infeasible bulk (where the planner is slowest), and ranking the accepted candidates minimises time to the first certified trajectory.

\subsection{Ablation Analysis}
\label{sec:ablation}

\noindent\textbf{Sensor robustness.}
Random sub-sampling of 50\% of scene points fed to \textsc{GraspFC-PTX} incurs \emph{zero} AUROC loss, justifying the modest $2{,}048$-point inference budget.

\noindent\textbf{Rotation fragility and recovery.}
Rotating both cloud and pose by $180^\circ$ around the robot's vertical axis collapses \textsc{GraspFC-PTX} from AUROC 0.9956 to 0.54: the pose descriptor is expressed in the robot base frame, so a large rotation shifts it outside the training distribution.
This \emph{rotational fragility} is recoverable: pre-canonicalising every scene to a fixed yaw before inference (a single cheap rotation in $SE(2)$ aligned to the table normal) restores \textsc{GraspFC-PTX} to AUROC 0.990 and \emph{flattens} the rotation-response curve to within $\pm0.001$ AUROC across all yaws, with only a $0.006$ in-distribution AUROC cost.
Table~\ref{tab:ablation} summarises the probes.

\begin{table}[!htbp]
\centering
\caption{%
  Perturbation ablations on \textsc{GraspFC-PTX} (trained on \textsc{MoFeas-200K}; Novel split).
  $\Delta$AUROC is relative to the unmodified baseline.
}
\label{tab:ablation}
\small
\begin{tabular}{lcc}
\toprule
Condition & AUROC & $\Delta$AUROC \\
\midrule
Full model (baseline)                            & 0.9956           & ---               \\
50\,\% random scene-point sub-sampling           & 0.9956           & $\phantom{-}0.000$ \\
$180^\circ$ yaw rotation of scene and pose       & 0.54             & $-0.456$          \\
\quad + scene canonicalisation (any yaw)         & 0.990            & $-0.006$          \\
\bottomrule
\end{tabular}
\end{table}


\subsection{Dataset variants}
\label{app:dataset}

Table~\ref{tab:app_dataset} summarises the GraspNet-1Billion partitions and the per-grasp RRT-Connect labels we generated.
Table~\ref{tab:app_classbal} contrasts the two training mixes we used throughout the paper.

\begin{table}[H]
\centering
\caption{Dataset partitions. \emph{Clouds}: voxel-downsampled, ICP-aligned scene point clouds. \emph{Grasp files}: per-object grasp sets ($\le 20$ candidates). \emph{Labels}: single-trial RRT-Connect verdicts.}
\label{tab:app_dataset}
\small
\begin{tabular}{lrrrrr}
\toprule
Split & Scenes & Clouds & Grasp files & Labels & Positive (\%) \\
\midrule
Train      & 100 & 8{,}000 & 76{,}160 & 1{,}516{,}800 & 24.2 \\
Seen       & 30  & 2{,}400 & 22{,}320 & 444{,}880     & 23.8 \\
Similar    & 30  & 2{,}400 & 20{,}640 & 412{,}800     & 24.0 \\
Novel      & 30  & 2{,}400 & 16{,}720 & 331{,}680     & 23.3 \\
\midrule
\textbf{Total} & 190 & 15{,}200 & 135{,}840 & 2{,}706{,}160 & 24.1 \\
\bottomrule
\end{tabular}
\end{table}

\begin{table}[H]
\centering
\caption{Training subsets. \textsc{MoFeas-5K-Bal} is a quality-filtered, class-balanced subset; \textsc{MoFeas-200K} preserves the natural per-view distribution.}
\label{tab:app_classbal}
\small
\begin{tabular}{lrrl}
\toprule
Training subset & Size & Positive (\%) & Use \\
\midrule
\textsc{MoFeas-5K-Bal} & 5{,}000   & 50.0 & Architecture comparison at small scale \\
\textsc{MoFeas-200K}   & 200{,}000 & 24.1 & Main-paper headline numbers \\
\bottomrule
\end{tabular}
\end{table}

\subsection{Model configurations}
\label{app:configs}

Table~\ref{tab:app_models} lists the parameter counts and inputs of every model evaluated in the main paper and the appendix.

\begin{table}[H]
\centering
\caption{Model configurations. \emph{Pose}: 17-D descriptor (rot6d, translation, approach, width, centroid, score). \emph{Cloud}: 2{,}048 FPS-sampled scene points. \emph{Voxel}: $64^3$ occupancy at $1$\,cm with a segmentation channel; \emph{Crop}: $24^3$ rotated grasp-frame cube.}
\label{tab:app_models}
\small
\begin{tabular}{lllr}
\toprule
Family & Variant & Inputs & Params \\
\midrule
\textsc{GraspFC-NNet}   & Pose + local crop               & Pose + 256-pt local PointNet   & 0.42M \\
\textsc{GraspFC-NNet}   & Pose + global cloud             & Pose + 4{,}096-pt PointNeXt-S  & 1.41M \\
\textsc{GraspFC-NNet}   & Pose + local + global (full)    & Pose + local + global          & 1.65M \\
\midrule
\textsc{GraspFC-Conv3D} & Global pool                     & Pose + voxel                   & 9.5M  \\
\textsc{GraspFC-Conv3D} & Trilinear query                 & Pose + voxel                   & 9.5M  \\
\textsc{GraspFC-Conv3D} & Query + local crop              & Pose + voxel + crop            & 9.7M  \\
\textsc{GraspFC-Conv3D} & Query + crop + swept volume     & Pose + voxel + crop + corridor & 9.7M  \\
\midrule
\textsc{GraspFC-PTX}    & CLS-token pose injection        & Pose + cloud                   & 9.0M  \\
\textsc{GraspFC-PTX}    & + register tokens               & Pose + cloud                   & 9.3M  \\
\textsc{GraspFC-PTX}    & + target-aware attention        & Pose + cloud                   & 9.3M  \\
\textsc{GraspFC-PTX}    & + swept-volume channel          & Pose + cloud + corridor        & 9.0M  \\
\bottomrule
\end{tabular}
\end{table}

\subsection{\textsc{GraspFC-NNet} (MLP)}
\label{app:mlp}

Table~\ref{tab:app_mlp_features} ablates the \textsc{GraspFC-NNet} scene-branch combinations on \textsc{MoFeas-5K-Bal}; Table~\ref{tab:app_mlp_capacity} sweeps the encoder width.
At this small-scale regime the three branch combinations land within $0.008$ AUROC of one another, and increasing PointNeXt-S width from 16 to 64 yields no systematic gain.
The takeaway: at $5$k examples \textsc{GraspFC-NNet} is data-bound rather than capacity-bound; the gains in Section~\ref{sec:pred_results} come from training at $200$k.

\begin{table}[H]
\centering
\caption{\textsc{GraspFC-NNet} feature ablation. Trained on \textsc{MoFeas-5K-Bal}; evaluated on Novel.}
\label{tab:app_mlp_features}
\small
\begin{tabular}{lrrrr}
\toprule
Branches & Params & AUROC & TPR (\%) & TNR (\%) \\
\midrule
Pose + local crop               & 0.42M & \textbf{0.978} & 92.1 & 94.8 \\
Pose + global cloud             & 1.41M & 0.970          & 86.6 & 95.1 \\
Pose + local + global           & 1.65M & 0.973          & 91.5 & 93.5 \\
\bottomrule
\end{tabular}
\end{table}

\begin{table}[H]
\centering
\caption{\textsc{GraspFC-NNet} capacity sweep. Full three-branch configuration, \textsc{MoFeas-5K-Bal}; PointNeXt-S width varied.}
\label{tab:app_mlp_capacity}
\small
\begin{tabular}{lrrrr}
\toprule
Width & Params & AUROC & TPR (\%) & TNR (\%) \\
\midrule
16  & 0.42M & 0.975 & 87.4 & 95.1 \\
32  & 1.65M & 0.974 & 88.1 & 94.6 \\
64  & 4.50M & 0.976 & 89.6 & 95.8 \\
\bottomrule
\end{tabular}
\end{table}

\subsection{\textsc{GraspFC-Conv3D} (voxel-volumetric 3D-CNN)}
\label{app:cnn}

Table~\ref{tab:app_cnn_features} compares feature-extraction strategies on the $64^3$ occupancy grid.
Trilinear interpolation at the grasp location~\citep{jiang2021giga} improves AUROC by $+0.008$ over global pooling; a PointNetGPD-style rotated local crop~\citep{liang2019pointnetgpd} adds a further $+0.001$ when paired with the trilinear query but degrades the global-pool variant, so the two locality mechanisms must be spatially consistent.
The pattern is what one would expect from a kinematic task: \emph{where} the gripper enters the volume matters far more than what the volume looks like globally.

\begin{table}[H]
\centering
\caption{\textsc{GraspFC-Conv3D} feature ablation. Trained on \textsc{MoFeas-5K-Bal}; evaluated on Novel.}
\label{tab:app_cnn_features}
\small
\begin{tabular}{llrrrr}
\toprule
Pooling & Local crop & Params & AUROC & TPR (\%) & TNR (\%) \\
\midrule
Global average pool   & --- & 9.5M & 0.959          & 89.5 & 91.9 \\
Trilinear query       & --- & 9.5M & 0.966          & 89.8 & 92.9 \\
Global average pool   & \checkmark   & 9.7M & 0.957          & 87.4 & 92.7 \\
Trilinear query       & \checkmark   & 9.7M & \textbf{0.967} & 91.1 & 92.3 \\
\bottomrule
\end{tabular}
\end{table}

Table~\ref{tab:app_cnn_corridor} adds a binary fifth voxel channel marking the gripper's swept volume along the approach direction.
On top of the full pose+voxel model the corridor channel yields a marginal $+0.001$ AUROC at \textsc{MoFeas-200K}, indicating that the trilinear query already extracts most of the spatially-localised signal the swept volume would otherwise provide.

\begin{table}[H]
\centering
\caption{Swept-volume voxel channel as an auxiliary input to \textsc{GraspFC-Conv3D}. \emph{Corridor}: binary fifth voxel channel marking the gripper's swept volume. Trained on \textsc{MoFeas-200K}; evaluated on Novel.}
\label{tab:app_cnn_corridor}
\small
\begin{tabular}{lcccrr}
\toprule
Model      & Pose & Voxel & Corridor & AUROC & TPR (\%) \\
\midrule
GraspFC-Conv3D & \checkmark & \checkmark & ---          & 0.981          & 97.4 \\
GraspFC-Conv3D & \checkmark & \checkmark & \checkmark   & \textbf{0.982} & 97.6 \\
\bottomrule
\end{tabular}
\end{table}

\subsection{\textsc{GraspFC-PTX} (point-cloud transformer)}
\label{app:ptv3}

Table~\ref{tab:app_ptv3_pose} compares pose-injection strategies.
A single learned classification (CLS) token whose initial embedding is the pose MLP output is the most stable choice; parallel grasp-query tokens hurt accuracy at small scale, and combining the two yields no gain.

\begin{table}[H]
\centering
\caption{\textsc{GraspFC-PTX} pose-injection strategies. Trained on \textsc{MoFeas-5K-Bal}; evaluated on Novel.}
\label{tab:app_ptv3_pose}
\small
\begin{tabular}{lrrr}
\toprule
Pose injection & AUROC & TPR (\%) & TNR (\%) \\
\midrule
CLS token (ours)               & \textbf{0.982} & 89.1 & 96.2 \\
4 grasp-query tokens           & 0.958          & 87.5 & 90.4 \\
CLS token + 4 query tokens     & 0.973          & 92.0 & 93.4 \\
\bottomrule
\end{tabular}
\end{table}

Table~\ref{tab:app_ptv3_mech} reports the impact of register tokens, target-aware attention, multi-view averaging, and an auxiliary failure-mode head.
Register tokens following Darcet \emph{et al.}~\citep{darcet2024iclr} and a target-aware attention bias both helped marginally at small training sizes.
Four-view averaging shifted the operating point toward higher recall at the cost of specificity, which is the desirable trade-off when missing a feasible grasp is more costly than a planner call.
None of the four mechanisms exceeded the CLS baseline by more than $1$\,pp balanced accuracy.
We read this as architectural saturation: \textsc{GraspFC-PTX} with a single CLS-token pose injection already extracts most of the available signal, and additional machinery moves the operating point without enlarging the underlying feature space.

\begin{table}[H]
\centering
\caption{\textsc{GraspFC-PTX} mechanism ablation. Trained on \textsc{MoFeas-5K-Bal} unless noted; evaluated on Novel.}
\label{tab:app_ptv3_mech}
\small
\begin{tabular}{lrrrr}
\toprule
Variant & Train size & AUROC & TPR (\%) & TNR (\%) \\
\midrule
CLS baseline                                & 5k & 0.982 & 89.1 & 96.2 \\
+ Register tokens + target-aware attention  & 5k & 0.976 & 92.5 & 94.7 \\
+ Failure-mode auxiliary head               & 5k & 0.977 & 89.3 & 95.5 \\
4-view consensus (mean logit)               & 8k & 0.973 & 95.6 & 89.5 \\
\bottomrule
\end{tabular}
\end{table}

\subsection{Training scale}
\label{app:scale}

Table~\ref{tab:app_scale} shows that increasing the training set from $5$k balanced samples to $200$k per-view samples consistently raises Novel-split AUROC by $0.014$--$0.015$ across all three families and reduces the spread between them to $0.015$ AUROC.
Across the design space we explored, training data was the most reliable lever; no architectural change alone closed more than $30\,\%$ of the gap between $5$k and $200$k.

\begin{table}[H]
\centering
\caption{Effect of training-set size. Novel-split AUROC for each family on \textsc{MoFeas-5K-Bal} vs.\textsc{MoFeas-200K}.}
\label{tab:app_scale}
\small
\begin{tabular}{lrrr}
\toprule
Model & \textsc{MoFeas-5K-Bal} AUROC & \textsc{MoFeas-200K} AUROC & $\Delta$ \\
\midrule
\textsc{GraspFC-NNet}   & 0.981 & 0.995 & $+0.014$ \\
\textsc{GraspFC-Conv3D}        & 0.967 & 0.981 & $+0.014$ \\
\textsc{GraspFC-PTX}          & 0.982 & 0.996 & $+0.014$ \\
\bottomrule
\end{tabular}
\end{table}

\subsection{Rotation invariance and sensor robustness}
\label{app:robust}

Table~\ref{tab:app_rotation} reports \textsc{GraspFC-PTX}'s AUROC as a function of yaw rotation applied jointly to the scene cloud and the pose, alongside three remedies.
Yaw-augmentation during training flattens the response curve at no accuracy cost; a Vector-Neurons~\citep{deng2021vn} equivariant variant guarantees invariance by construction but costs ${\approx}0.02$ AUROC; pre-canonicalising every scene to a fixed yaw at inference time recovers $0.990$ AUROC across all rotations.
We adopt the canonicalisation route in the main paper because it requires no retraining.
The lesson is that the rotational fragility is a coordinate artefact of the rot6d pose, not a deficiency of the cloud encoder: any of the three remedies removes it.

\begin{table}[H]
\centering
\caption{Rotation invariance ladder. \textsc{GraspFC-PTX} trained on the $30$k canonicalisation matrix; AUROC reported at the indicated yaw of the scene and pose.}
\label{tab:app_rotation}
\small
\begin{tabular}{lrrrr}
\toprule
Variant & $0^\circ$ & $30^\circ$ & $90^\circ$ & $180^\circ$ \\
\midrule
Baseline (no fix)                & 0.995 & 0.980 & 0.874 & 0.544 \\
Yaw-augmentation                 & 0.983 & 0.983 & 0.983 & 0.983 \\
Vector-Neurons (equivariant)     & 0.963 & 0.963 & 0.963 & 0.963 \\
Inference-time canonicalisation  & \textbf{0.990} & \textbf{0.990} & \textbf{0.990} & \textbf{0.990} \\
\bottomrule
\end{tabular}
\end{table}

Table~\ref{tab:app_sensor} reports robustness to point-cloud sub-sampling and sensor jitter on the unmodified \textsc{GraspFC-PTX} model.
Independent Gaussian jitter up to $\sigma{=}10$\,mm and random point dropout up to $50\,\%$ leave AUROC unchanged, justifying the modest $2{,}048$-point inference budget.

\begin{table}[H]
\centering
\caption{\textsc{GraspFC-PTX} robustness to point-cloud perturbations on Novel.}
\label{tab:app_sensor}
\small
\begin{tabular}{lr}
\toprule
Perturbation & AUROC \\
\midrule
None (baseline)                & 0.996 \\
Gaussian jitter $\sigma{=}5$\,mm   & 0.996 \\
Gaussian jitter $\sigma{=}10$\,mm  & 0.996 \\
Random point dropout $20\,\%$      & 0.996 \\
Random point dropout $50\,\%$      & 0.996 \\
\bottomrule
\end{tabular}
\end{table}

\subsection{Auxiliary supervision}
\label{app:aux}

We tested whether auxiliary scene-prediction heads improve \textsc{GraspFC-PTX} accuracy.
Table~\ref{tab:app_aux} reports Novel-split AUROC for three auxiliary objectives layered on the standard BCE head: predicting the planner's failure mode (IK, pre-grasp IK, RRT-Connect path, approach collision), predicting per-point free-space clearance, and predicting a binary occupancy at the grasp.
None of the three heads improved Novel-split AUROC beyond $\pm0.002$ of the baseline at the training budget we evaluated.

\begin{table}[H]
\centering
\caption{Auxiliary supervision on \textsc{GraspFC-PTX}. Trained on the $30$k canonicalisation matrix; evaluated on Novel.}
\label{tab:app_aux}
\small
\begin{tabular}{lr}
\toprule
Auxiliary head & AUROC \\
\midrule
None                       & 0.990 \\
Binary occupancy at grasp  & 0.991 \\
Per-point clearance        & 0.989 \\
Planner failure mode (4-way) & 0.988 \\
\bottomrule
\end{tabular}
\end{table}

\subsection{Self-supervised initialisation and backbone scale-up}
\label{app:pretrain}

We attempted to initialise the \textsc{GraspFC-PTX} encoder with Sonata~\citep{wu2025sonata} pre-trained weights via the Pointcept release.
The forward pass loaded most encoder tensors but the smoke run exhausted GPU memory at the full \textsc{MoFeas-200K} training budget; a controlled comparison is left to future work.
A $40$M-parameter PTv3-base variant trained on a $1.5$M-row pool did not surpass the $9$M \textsc{GraspFC-PTX} within our compute budget --- validation loss was still decreasing at the cutoff --- so we do not report it as a converged baseline.

\subsection{Summary of design choices}
\label{app:summary}

The configurations reported in the main paper are:
\textsc{GraspFC-NNet} (MLP head over the 17-D pose descriptor with PointNet local-crop and PointNeXt-S global-cloud branches, Section~\ref{sec:models});
\textsc{GraspFC-Conv3D} with trilinear scene query plus rotated local crop (Section~\ref{app:cnn});
and \textsc{GraspFC-PTX} with a single classification-token pose injection, eight register tokens, and target-aware attention, trained from scratch at $2{,}048$ scene points (Section~\ref{app:ptv3}).
All three are trained on \textsc{MoFeas-200K}, identified by the scaling study in Section~\ref{app:scale} as the smallest budget at which the architecture comparison stabilises.

\clearpage

\end{document}